\documentclass[sigconf]{acmart}
\renewcommand\footnotetextcopyrightpermission[1]{} 
\AtBeginDocument{%
  \providecommand\BibTeX{{%
    \normalfont B\kern-0.5em{\scshape i\kern-0.25em b}\kern-0.8em\TeX}}}

\setcopyright{none}

\acmSubmissionID{5}

\usepackage{footnote}
\begin{document}
\sloppy
\title{Deep Contextual Embeddings for Address Classification in E-commerce}

\author{Shreyas Mangalgi}
\email{shreyas.mangalgi@myntra.com}
\affiliation{%
 \institution{Myntra Designs Pvt. Ltd.\\India}
}

\author{Lakshya Kumar}
\email{lakshya.kumar@myntra.com}
\affiliation{%
 \institution{Myntra Designs Pvt. Ltd.\\India}
 }


\author{Ravindra Babu T}
\email{ravindra.babu@myntra.com}
\affiliation{%
 \institution{Myntra Designs Pvt. Ltd.\\India}
}

\renewcommand{\shortauthors}{Mangalgi and Lakshya et al.}

\begin{abstract}
E-commerce customers in developing nations like India tend to follow no fixed format while entering shipping addresses. Parsing such addresses is challenging because of a lack of inherent structure or hierarchy. It is imperative to understand the language of addresses, so that shipments can be routed without delays. In this paper, we propose a novel approach towards understanding customer addresses by deriving motivation from recent advances in Natural Language Processing (NLP). We also formulate different pre-processing steps for addresses using a combination of edit distance and phonetic algorithms. Then we approach the task of creating vector representations for addresses using Word2Vec with TF-IDF, Bi-LSTM and BERT based approaches. We compare these approaches with respect to sub-region classification task for North and South Indian cities. Through experiments, we demonstrate the effectiveness of generalized RoBERTa model, pre-trained over a large address corpus for language modelling task. Our proposed RoBERTa model achieves a classification accuracy of around \textbf{90\%} with minimal text preprocessing for sub-region classification task outperforming all other approaches. Once pre-trained, the RoBERTa model can be fine-tuned for various downstream tasks in supply chain like pincode \footnote{Equivalent to zipcode} suggestion and geo-coding. The model generalizes well for such tasks even with limited labelled data. To the best of our knowledge, this is the first of its kind research proposing a novel approach of understanding customer addresses in e-commerce domain by pre-training language models and fine-tuning them for different purposes.
\end{abstract}



\maketitle

\section{Introduction}
Machine processing of manually entered addresses poses a challenge in developing countries because of a lack of standardized format. Customers shopping online tend to enter shipping addresses with their own notion of correctness. This creates problems for e-commerce companies in routing shipments for last mile delivery.
Consider the following examples of addresses entered by customers:
\begin{enumerate} 
    \item \textit{`XXX\footnote{We do not mention exact addresses to protect the privacy of our customers}, AECS Layout, Geddalahalli, Sanjaynagar main Road Opp. Indian Oil petrol pump, Ramkrishna Layout, Bengaluru Karnataka 560037'}\\
    
    \item \textit{`XXX, B-Block, New Chandra CHS, Veera Desai Rd, Azad Nagar 2, Jeevan Nagar, Azad Nagar, Andheri West, Mumbai, Maharashtra 400102'} \\
    
    \item \textit{`Gopalpur Gali XXX, Near Hanuman Temple, Vijayapura, Karnataka 586104'} \\
    
     \item \textit{`Sector 23, House number XXX, Faridabad, Haryana 121004'} \\
    \item \textit{`H-XXX, Fortune Residency, Raj Nagar Extension Ghaziabad Uttar Pradesh 201003'}
\end{enumerate}
It is evident from above illustrations that addresses do not tend to follow any fixed pattern and consist of tokens with no standard spellings. Thus, applying Named Entity Recognition (NER) systems to Indian addresses for sub-region classification becomes a challenging problem. Devising such a system for Indian context requires a large labelled dataset to cover all patterns across the geography of a country and is a tedious task. At the same time, Geo-location information which otherwise makes the problem of sub-region classification trivial, is either not readily available or is expensive to obtain. In spite of all these challenges, e-commerce companies need to deliver shipments at customer doorstep in remote as well as densely populated areas. At this point, it becomes necessary to interpret and understand the language of noisy addresses at scale and route the shipments appropriately. Many a times, fraudsters tend to enter junk addresses and e-commerce players end up incurring unnecessary shipping and reverse logistic costs. Hence, it is important to flag incomplete addresses while not being too strict on the definition of completeness. In recent years, the focus of NLP research has been on pre-training language models over large datasets and fine-tuning them for specific tasks like text classification, machine translation, question answering etc.
In this paper, we propose methods to pre-process addresses and learn their latent representations using different approaches. Starting from traditional Machine Learning method, we explore sequential network\cite{lstm_hochreiter} and Transformer\cite{transformers} based model to generate address representations. We compare these different paradigms by demonstrating their performance over sub-region classification task. We also comment on the limitations of traditional Machine Learning approaches and advantages of sequential networks over them. Further, we talk about the novelty of Transformer based models over sequential networks in the context of addresses. 
The contribution of the paper is as follows:
\begin{enumerate}
    \item Details the purpose and challenges of parsing noisy addresses
    \item Introduces multi-stage address preprocessing techniques
    \item Proposes three approaches to learn address representations and their comparison with respect to sub-region classification task 
    \item Describes advantages of BERT based model as compared to traditional methods and sequential networks in the context of Indian addresses
\end{enumerate}
The rest of the paper is organized as follows: In Section \ref{sec:related work}, we review previous works that deal with addresses in e-commerce. Next we present insights into the problem that occur in natural language addresses in Section \ref{sec:problem Insights} and propose pre-processing steps for addresses in Section \ref{sec: Address Pre-processing}. In Section \ref{sec:approaches}, we present different approaches to learn latent address representations with sub-region classification task. In Section \ref{sec:experimental setup}, we outline the experimental setup and present the results and visualizations of our experiments in Section \ref{sec:results}. We present error analysis in Section \ref{sec:error analysis} where we try to explain the reasons behind misclassified instances. Finally, we conclude the paper and discuss future work in Section \ref{sec:conclusion}.

\section{Related Work}
\label{sec:related work}
\citet{10.1145/2047296.2047297} provide a detailed account of seven major issues in Geographical Information Retrieval such as detecting and disambiguation of toponyms, geographical relevant ranking and user interfaces. Among them, the challenges of vague geographic terminology and spatial and text indexing are present in the current problem. Some such frequently occurring spatial language terms in the addresses include `near', `behind' and `above'. \citet{Linet} propose hierarchical cluster and route procedure to coordinate vehicles for large scale post-disaster distribution and evaluation. \citet{RePEc:elg:eechap:14398_3} provide a description of the last mile problem in logistics and the challenges uniquely faced by last mile unlike any other component in supply chain logistics. \citet{DBLP:conf/gir/BabuCKSG15} show methods to classify textual address into geographical sub-regions for shipment delivery using Machine Learning. Their work mainly involves address pre-processing, clustering and classification using ensemble of classifiers to classify the addresses into sub-regions. \citet{Loke} work on Malaysian addresses and classify them into different property types like condominium, apartments, residential homes ( bungalow, terrace houses, etc.) and business premises like shops, factories, hospitals, and so on. They use Machine Learning based models and also propose LSTM based models for classifying addresses into property types. \citet{Kakkar2018AddressCF} discuss the challenges with address data in Indian context and propose methods for efficient large scale address clustering using conventional as well as deep learning techniques. They experiment with different variants of Leader clustering using edit distance, word embedding etc. For detecting fraud addresses over e-commerce platforms and reduce operational cost, \citet{Babu2017AddressFM} propose different Machine Learning methodologies to classify addresses as `normal' or `monkey-typed'(fraud). In order to represent addresses of populated places, \citet{10.1007/978-3-319-68204-4_14} train a neural embedding algorithm based on skip-gram\cite{mikolov:2013} architecture to represent each populated place into a 100-dimensional vector space. For Neural Machine Translation (NMT), \citet{transformers} propose Transformer model which is based solely on attention mechanisms, moving away from recurrence and convolutions. The transformer model significantly reduces training time due to its positional embeddings based parallel input architecture. \citet{BERT} propose BERT architecture, which stands for Bidirectional Encoder Representations from Transformers. They demonstrate that BERT obtains state-of-the-art performance on several NLP tasks like question answering, classification etc. \citet{Roberta} propose RoBERTa which is a robust and optimized version of pre-training a BERT based model and achieve new state-of-the-art results on GLUE\cite{wang-etal-2018-glue}, RACE\cite{lai-etal-2017-race} and SQuAD\cite{DBLP:journals/corr/RajpurkarZLL16} datasets. \newline
To the best of our knowledge, there is no prior work that treats the problem of understanding addresses from a language modelling perspective. We experiment with different paradigms and demonstrate the efficacy of a BERT based model, pre-trained over a large address corpus for the downstream task of sub-region classification.

\section{Problem Insights}\label{sec:problem Insights}
\begin{figure}[h]
  \centering
  \includegraphics[width=\columnwidth]{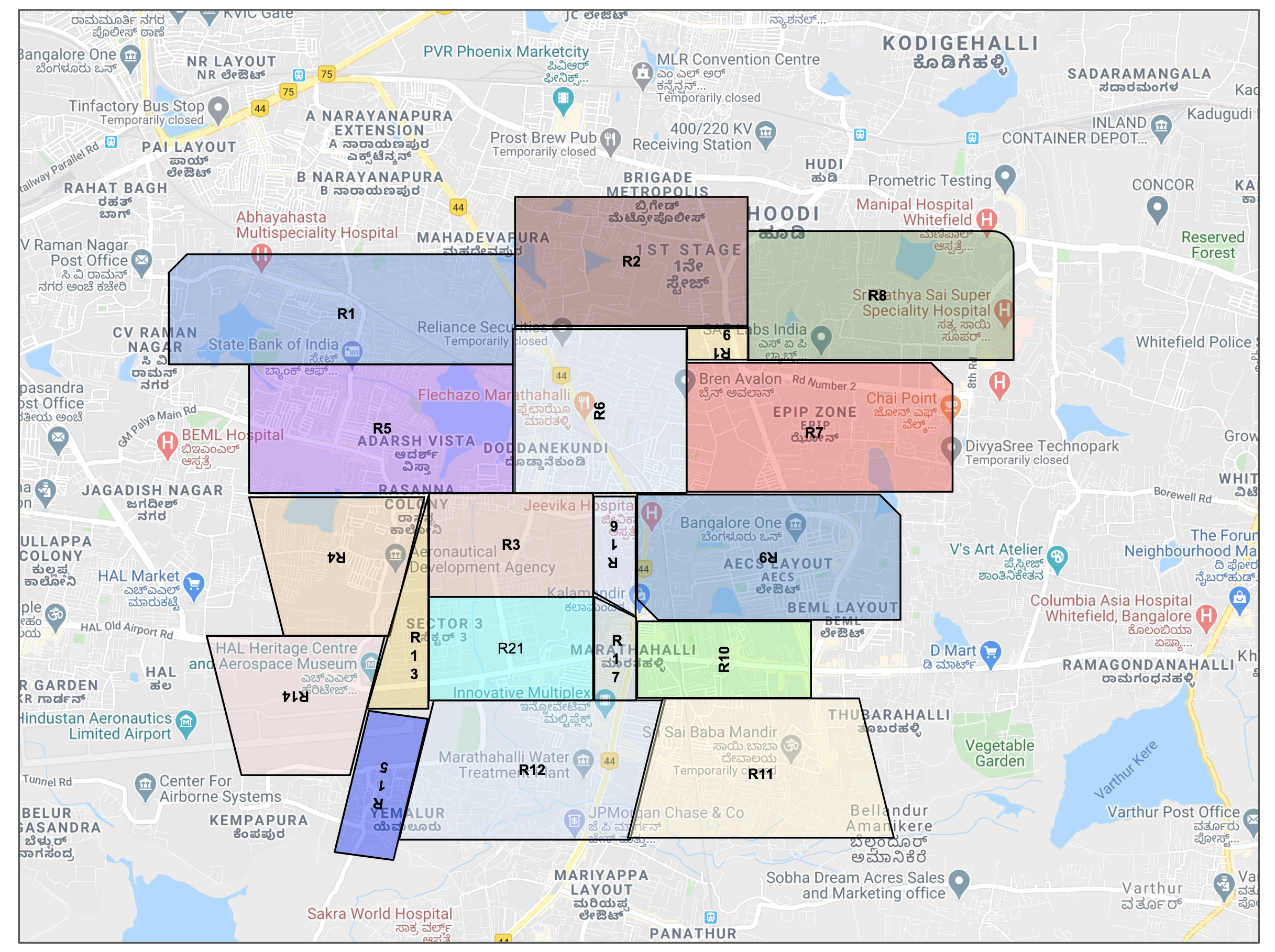}
  \caption{Sample map depicting sub-regions for last mile delivery.}
  \label{fig:last mile delivery map}
\end{figure}

Sorting shipments based on addresses forms an integral part of e-commerce operations. When a customer places an order online, items corresponding to the order are packed at a warehouse and dispatched for delivery. During the course of its journey from warehouse to the customer doorstep, the shipment undergoes multiple levels of sortation. The first level of sort is very broad and typically happens at the warehouse itself where multiple shipments are clubbed together in a `master bag' on a state level and dispatched to different hubs. At the hub, another sort takes place on a city level and shipments are dispatched to the respective cities. As customers usually provide their state and city information accurately, sorting at the warehouse and the hub becomes a trivial process. Once shipments reach a city, they are sub-divided into different zones. Each zone is further divided into multiple sub-regions. The sub-regions can take highly irregular shapes depending on density of customers, road network and ease of delivery. Figure \ref{fig:last mile delivery map} shows different sub-regions for last mile delivery of shipments\footnote{photograph is captured from Google maps}. These sub-regions constitute the class definitions for our problem. The challenge in last mile delivery emerges when customers are unsure of their locality names and pincodes. The ambiguity arises because of the unstructured nature of localities and streets in developing nations. Coupled with this, area names originating from colloquial languages make it difficult for users to enter their shipping addresses in English, resulting in multiple spell variants of localities, sometimes in hundreds. Solutions for parsing addresses \cite{10.1145/3347146.3359070} have been proposed in the past but they seldom work for cities in developing nations like India, Nepal and Bangladesh where there is no standard way of writing an address. The notion of sufficiency of information for a successful delivery is subjective and depends on multiple factors like the address text, familiarity of locality for the delivery agent, availability of customers' phone numbers in case of confusion and so on. In planned cities, areas are generally divided into blocks (sectors) and thus mentioning the house number and name of the sector might be sufficient for a delivery agent. But in unplanned cities, where a huge majority of the population resides, locality definition become very subjective. Deciphering pincodes is also a cumbersome task, especially at the boundaries. As a result, customers end up entering noisy addresses for the last mile. For the purpose of the problem described in this paper, we ignore phone numbers of customers as a source of information and work with text addresses only. For solving the challenges of last mile delivery, we propose address preprocessing methods. Subsequently, we describe ways to use state-of-the-art NLP approaches to obtain address representations which can be used for various downstream tasks.

\section{Address Pre-processing}\label{sec: Address Pre-processing}
\begin{figure}[h]
  \centering
  \includegraphics[scale = 0.15]{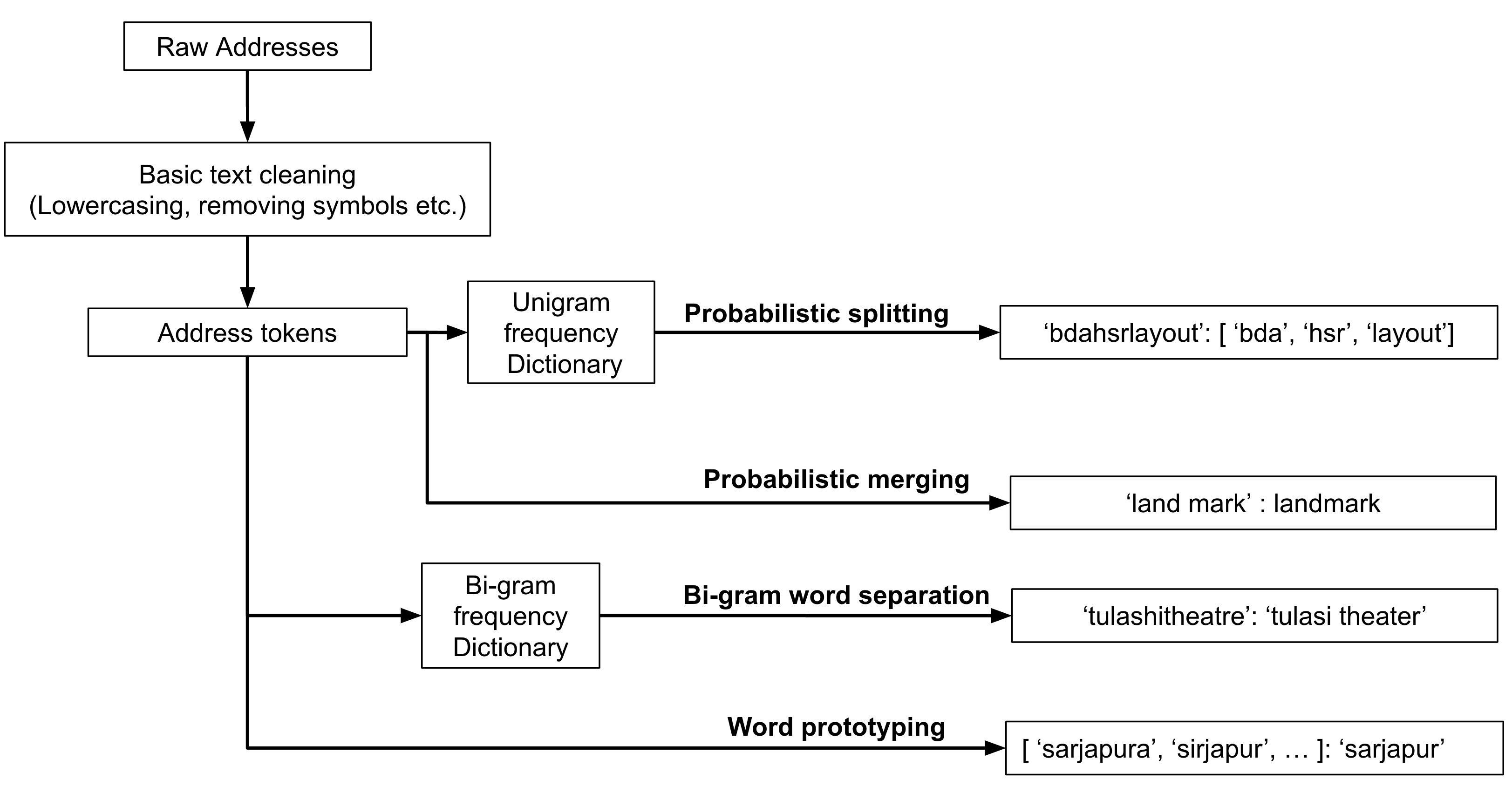}
  \caption{Flow-Diagram depicting address pre-processing methods.}
  \Description{}
  \label{fig:address_cleaning}
\end{figure}

We propose different pre-processing steps for addresses. Our analysis indicates that customers generally tend to make mistakes while entering names of localities and buildings broadly due to two reasons: 

\begin{enumerate}
    \item Input errors while typing, arising mainly due to closeness of characters in the keyboard
    \item Uncertainty regarding `correct' spellings of locality/street names
\end{enumerate}
The notion of `correct spelling' is itself subjective and we consider the correct spelling to be the one entered most frequently by customers. In our work, we bucket the errors into four categories in Table \ref{table:error categories}. The second column shows address tokens and their correct form.  
\begin{table}[!h]
\centering
  \resizebox{\columnwidth}{!}{
   \begin{tabular}{|c|c|c|}\hline
     \textbf{Error Type} & \textbf{Example} \\ \hline
   
    Missing white space between correctly spelled tokens & \textit{meenakshiclassic} $\rightarrow$ \textit{meenakshi classic} \\ \hline 
    
    Redundant whitespace between correctly spelled tokens &  \textit{lay out} $\rightarrow$ \textit{layout} \\ \hline
  
    Misspelled individual tokens &  \textit{appartments} $\rightarrow$ \textit{apartments} \\ \hline
    
    Misspelled compound tokens with no whitespace &  \textit{sectarnoida} $\rightarrow$ \textit{sector noida} \\ \hline 
  \end{tabular}}
\caption{Types of errors encountered in Indian Addresses.}
\label{table:error categories}
\end{table}
For traditional ML approaches, directly using raw addresses without spell correction leads to a larger vocabulary size bringing in problems of high dimensionality and over-fitting. Standard English language Stemmers and Lemmatizers do not yield satisfactory results for vocabulary reduction because of code-mixing \cite{1510da41515d46a9866d1087b5c8f052} issues in addresses. Hence we need to devise custom methods for reducing the vocabulary size by correcting for spell variations. We illustrate different methods for solving each of the errors mentioned in Table \ref{table:error categories}. Figure \ref{fig:address_cleaning} shows the overall steps involved in address preprocessing.

\subsection{\textbf{Basic Cleaning}}\label{sec:basic cleaning}
We begin by performing a basic pre-processing of addresses which includes removal of special characters and lower-casing of tokens. We remove all the numbers which are of length greater than six\footnote{In India pincodes are of length six} as some customers enter phone numbers and pincodes in the address field. Address tokens are generated by splitting on whitespace character. We append pincodes to addresses after applying all the pre-processing steps.

\subsection{\textbf{Probabilistic splitting}}\label{sec:Probabilistic splitting}
The motivation for this technique comes from \citet{DBLP:conf/gir/BabuCKSG15}. In this step, we perform token separation using count frequencies in a large corpus\footnote{We refer to `corpus' and `dataset' synonymously in this paper} of customer addresses. The first step is to construct a term-frequency dictionary for the corpus. Following this, we iterate through the corpus and split each token at different positions to check if the resulting count of individual tokens after splitting is greater than the compound token. In such cases, we store the instance in a separate dictionary which we can use for preprocessing addresses at runtime. Consider the compound token \textbf{\textit{`hsrlayout'}}, the method iteratively splits this token at different positions and finally results in \textbf{\textit{`hsr'}} and \textbf{\textit{`layout'}} as the joint probability of these tokens exceeds the probability of \textbf{\textit{`hsrlayout'}}.



\subsection{\textbf{Spell correction}}\label{sec:Word prototyping}
In order to determine the right spell variant and correct for variations, we cluster tokens in our corpus using leader clustering \cite{doi:10.1080/00401706.1981.10487693}. This choice emanates from the fact that leader clustering is easy to implement and does not require specifying the number of clusters in advance and has a complexity of $O(n)$. We recursively cluster the tokens in our corpus using Levenshtein distance \cite{1966SPhD...10..707L} coupled with Metaphone algorithm \cite{clm/philips90}. We use a combination of these algorithms as standalone use of either of them has some drawbacks. While using only edit distance based clustering, we observe that many localities which differ by a single character but are phonetically different tend to get clustered together. For example, in a city like Bangalore we find two distinct localities with names `\textit{\textbf{Bommasandra}}' and `\textit{\textbf{Dommasandra}}' which differ by a single character. While using only the edit distance condition, all instances of the former will be tagged with the latter or vice-versa depending on which locality occurs more frequently in the corpus. This will result in erroneous outputs for downstream tasks. In the above instance, we observe that the two locality names are phonetically different and a phonetic algorithm will output different hash values and thus help in assigning them to separate clusters. When using only phonetic based approach, we find that some distinct locality names are clustered together. For example, two localities by the name `\textit{\textbf{Mathkur}}' and `\textit{\textbf{Mathikere}}' have the same hash value according to Metaphone algorithm. In this case, even though the two locality names have similar phonetic hash values, by using an edit distance threshold we can ensure that the two are not clustered together. This inspires our choice of a combination of edit distance and phonetic algorithms for clustering spell variants. The most frequently occurring instance within each cluster is selected as the `leader'. In order to perform spell correction, we replace each spell variant with its corresponding leader.
The key principle behind spell correction for addresses is summarized below.
Consider two tokens $\boldsymbol{T_{a}}$ and  $\boldsymbol{T_{b}}$. We say $\boldsymbol{T_{a}}$ is a spell variant of $\boldsymbol{T_{b}}$ if:
\begin{center}
  $Count(\boldsymbol{T_{a}}) < Count(\boldsymbol{T_{b}})$ 
\\ \textbf{\&}
\\ $Levenshtein Distance(\boldsymbol{T_{a}},\boldsymbol{T_{b}}) < threshold$ 
\\ \textbf{\&}
\\ $Metaphone(\boldsymbol{T_{a}}) == Metaphone(\boldsymbol{T_{b}})$
\end{center}
For experiments, we set $threshold$ value equal to $\textbf{3}$ and only consider tokens of length greater than $\textbf{6}$ as candidates for spell correction.
\footnote{We set the edit distance threshold and minimum token length values after performing empirical studies}

\subsection{\textbf{Bigram separation}}
\label{sec:Bigram separation}
Probabilistic splitting of compound tokens described in Section \ref{sec:Probabilistic splitting} can work only when individual tokens after splitting have a support larger than the compound token. If there is a missing white space coupled with a spelling error, probabilistic word splitting will not be able to separate the tokens. This is because the incorrect spell variant will not have enough support in the corpus. For this reason, we propose bi-gram word separation using leader clustering. At first, we construct a dictionary of all bi-grams occurring in our corpus as keys and number of occurrences as values. These bigrams are considered as single tokens and we iterate through the corpus and cluster the address tokens using leader clustering algorithm with edit distance threshold and phonetic matching condition. For example, \textbf{\textit{`bangalore karnataka'}} is a bigram which occurs frequently in the corpus. Against this `leader', different erroneous instances get assigned like \textbf{\textit{`bangalorkarnataka'}}, \textbf{\textit{`bangalorekarnatak'}} etc. These instances cannot be split using `Probabilistic splitting' described in Section \ref{sec:Probabilistic splitting} since the tokens \textbf{\textit{`bangalor'}} and \textbf{\textit{`karnatak'}} do not have significant support in the corpus. We store the bigrams and their error variants in a dictionary and use it for address pre-processing in the downstream tasks.

\subsection{\textbf{Probabilistic merging}}
\label{sec:Probabilistic merging}
Customers often enter unnecessary whitespaces while typing addresses. For correcting such instances, we propose probabilistic merging similar to the method described in Section \ref{sec:Probabilistic splitting}. Instead of splitting the tokens, we merge adjacent tokens in the corpus if the compound token has a higher probability of occurrence compared to the individual tokens. For example, the tokens \textbf{\textit{`lay'}} and \textbf{\textit{`out'}} will have a significantly lesser count than the compound token \textbf{\textit{`layout'}} and thus all such instances where the token \textit{\textbf{`lay'}} occurs followed by \textbf{\textit{`out'}} will be replaced with \textbf{\textit{`layout'}}.

\section{Approaches}\label{sec:approaches}
We demonstrate the use of three different paradigms: Traditional Machine Learning, Bi-LSTMs and BERT based model for generating latent address representations and use them for sub-region classification task. Through our experiments, we observe the effectiveness of generalized RoBERTa model, pre-trained over a large address corpus for language modelling task. We also comment on the limitations of traditional machine learning approaches and advantages of sequential networks. Further, we talk about the novelty of RoBERTa model over sequential networks in the context of addresses.

\subsection{Word2Vec with TF-IDF}
\label{sec:tf_idf}
As a baseline approach we use the techniques described in Section \ref{sec: Address Pre-processing} to pre-process addresses and use them to train a Word2Vec model \cite{mikolov:2013} for obtaining vector representation of tokens in an address. Further we compute Term Frequency - Inverse Document Frequency (TF-IDF)\cite{salton1986introduction} values for each token within an address and use them as weights for averaging the word vectors to obtain representation for an address. Weighting the tokens within an address is necessary since not all tokens in an address are equally `important' from a classification standpoint. Generally, we find that locality/landmark is the most important token, even for humans to identify the right sub-region. But in many instances, it is not very straight forward to be able to point at the appropriate information necessary for classification. Thus, in order to make the model automatically identify the most important tokens in an address, we use the TF-IDF concept. TF-IDF is a statistical measure of how important a word is to a document in a collection or corpus. Term Frequency (TF) denotes the frequency of a term within a document and Document Frequency (DF) is a measure of the number of documents in which a particular term occurs. Mathematically, it can be defined as follows:
\begin{gather}
\textbf{TF}(t,d)= f_{t,d}
\end{gather}
\begin{gather}
\textbf{IDF}(t,D)= log\frac{N}{|\{ d \in D: t \in d \}|}
\end{gather}
In our case, an address is a document $\boldsymbol{d}$, each token in the address is $\boldsymbol{t}$, the collection of addresses is the corpus $\boldsymbol{D}$ and total number of addresses is $\boldsymbol{N}$.
The TF-IDF statistic by definition assigns a lower weight to very frequent tokens. For our case, city names are typically mentioned in all the addresses and thus are not of much value. Contrary to city names, customers sometime enter information like detailed directions to reach their door step which form rare tokens in our corpus. In such cases, TF-IDF assigns the maximum weight of $log(N)$ and we remove such tokens while constructing address representations. We use these embeddings as features for training a multi-class classifier to classify addresses into sub-regions. A drawback of this approach is that, by averaging word vectors we end up losing the sequential information. For example, the addresses \textbf{`House No. X, Sector Y, Faridabad'} and \textbf{`House No. Y, Sector X, Faridabad'}\footnote{X, Y here are typically numerical values} end up producing the same address vectors and induce error in classification. In spite of this, it forms a strong baseline to assess the efficacy of more advanced approaches.

\begin{figure}[h]
  \centering
  \includegraphics[scale = 0.30]{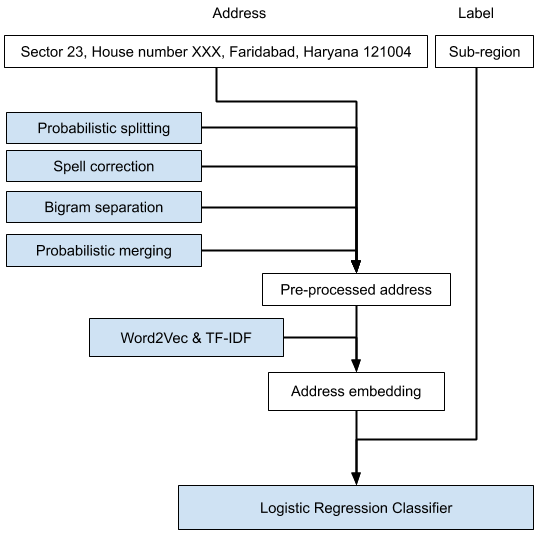}
  \caption{Logistic regression classifier with TF-IDF weighted Word2Vec vectors as features.}
  \Description{}
\end{figure}

\subsection{Bi-LSTM}\label{sec:bilstm}
Averaging word vectors leads to loss of sequential information, hence we move on to a Bi-LSTM\cite{lstm_hochreiter} based approach. While LSTMs are known to preserve the sequence information, Bi-directional LSTMs have an added advantage since they also capture both the left and right context in case of text classification. Given an input address of length \textit{T} with words $w_{t}$, where $t \in [1, T]$. We convert each word $w_{t}$ to its vector representation $x_{t}$ using the embedding matrix $E$. We then use a Bi-directional LSTM to get annotations of words by summarizing information from both directions. Bidirectional LSTMs consist of a forward LSTM $ \overrightarrow{f}$, which reads the address from $ w_{1} $ to $ w_{T} $ and a backward LSTM $ \overleftarrow{f}$, which reads the address from $ w_{T} $ to $ w_{1} $:
    \begin{gather}
    	x_{t} = E^{T}w_{t},    t \in [1,T] \\
    	\overrightarrow{h_{t}} = \overrightarrow{LSTM}(x_{t}),    t \in [1,T] \\
    	\overleftarrow{h_{t}} = \overleftarrow{LSTM}(x_{t}),    t \in [T,1]
    \end{gather}

We obtain representation for a given address token $w_{t}$ by concatenating the forward hidden state $\overrightarrow{h_{t}}$ and backward hidden state $\overleftarrow{h_{t}}$, i.e., $h_{t} = [\overrightarrow{h_{t}}, \overleftarrow{h_{t}}]$, which summarizes the information of the entire address centered around $w_{t}$. The concatenation of final hidden state outputs of the forward and backward LSTMs is denoted by $h_{T} = [\overrightarrow{h_{T}}, \overleftarrow{h_{T}}]$. This forms the address embedding which is then passed to a dense layer with softmax activation. We train this model by minimizing cross-entropy loss. Section \ref{sec:results} shows the improvement in performance due to the ability of LSTMs to better capture sequential information. The drawback with this approach is that training a Bi-LSTM model is relatively slow because of its sequential nature. Hence we experiment with Transformer based models which have parallelism in-built. 


\begin{figure}[h]
  \centering
  \includegraphics[scale = 0.29]{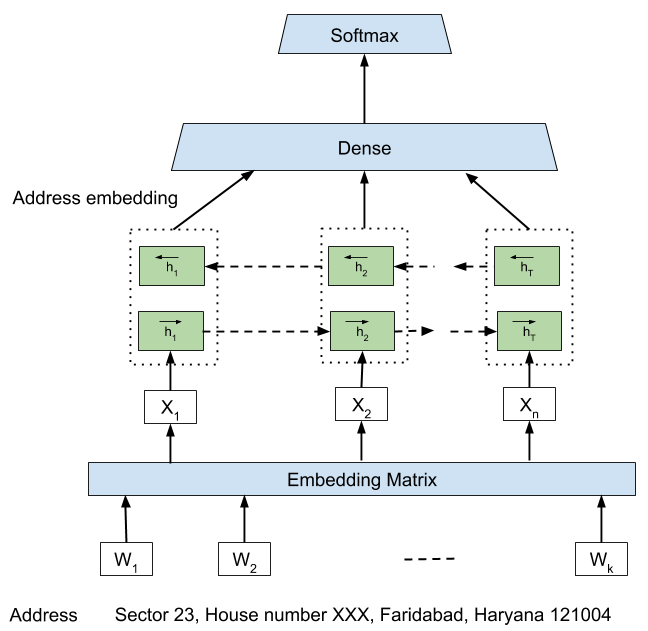}
  \caption{Bi-LSTM architecture for Address classification.}
  \Description{}
\end{figure}

\subsection{RoBERTa}\label{sec: BERT}
\begin{figure}[h]
  \centering
  \includegraphics[scale = 0.29]{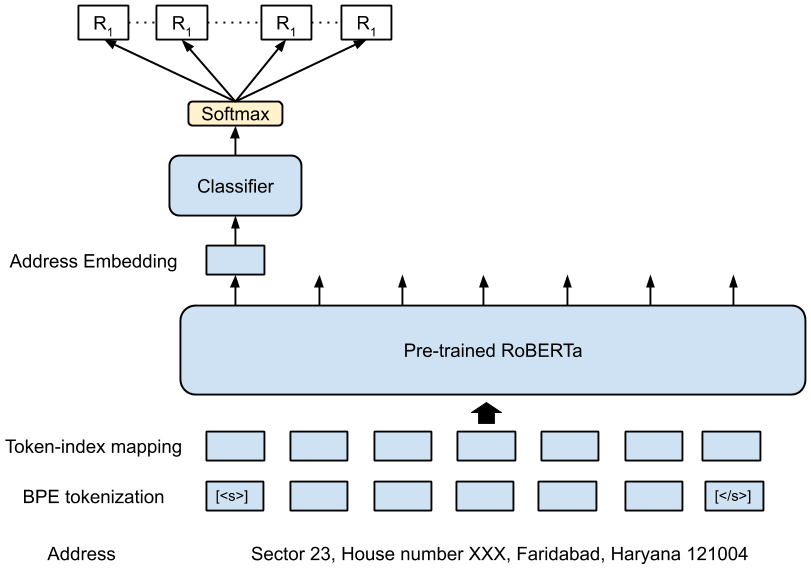}
  \caption{Pre-trained RoBERTa architecture for sub-region classification.}
  \Description{}
  \label{fig:BERT Model}
\end{figure}

In this section, we experiment with RoBERTa\cite{Roberta} which is a variant of BERT\cite{BERT}, for pre-training over addresses and fine-tune it for sub-region classification task. The BERT model optimizes over two auxiliary pre-training tasks: 
 \begin{itemize}
     \item \textbf{Mask Language Model (MLM)}: Randomly masking 15\% of the tokens in each sequence and predicting the missing words
     \item \textbf{Next Sentence Prediction (NSP)}: Randomly sampling sentence pairs and predicting whether the latter sentence is the next sentence of the former
 \end{itemize}
 BERT based representations try to learn the context around a word and is able to better capture its meaning syntactically and semantically. For our context, NSP loss does not hold meaning since customer addresses on e-commerce platforms are logged independently. This motivates the choice of RoBERTa model since it uses only the MLM auxiliary task for pre-training over addresses. In experiments we use byte-level BPE \cite{BPE} tokenization for encoding addresses. We use perplexity\cite{chen_beeferman_rosenfeld_2018} score for evaluating the RoBERTa language model. It is defined as follows: 
 \begin{gather}
    	P(Sentence) = P(w_{1}w_{2}....w_{N})^{\frac{-1}{N}}     
\end{gather}
where P(Sentence) denotes the probability of a test sentence and $w_{1}$, $w_{2}$,...., $w_{N}$ denotes words in the sentence. Generally, lower is the perplexity, better is the language model. After pre-training, the model would have learnt the syntactic and semantic aspects of tokens in shipping addresses. Figure \ref{fig:BERT Model} shows the overall approach used to pre-train RoBERTa model and fine-tune it for sub-region classification.

\begin{table}[!h]
\centering
  \resizebox{\columnwidth}{!}{%
   \begin{tabular}{|c|c|c|c|}\hline
     \textbf{Dataset} & \textbf{Number of rows (addresses)} & \textbf{Number of classes (sub-regions)} \\ \hline
   
    Zone-1 & 98,868 & 42 \\ \hline
    
    Zone-2 & 93,991 &  20 \\ \hline
    
    Zone-3 & 106,421 & 86 \\ \hline 
    
    Zone-4 & 218,434 & 176 \\ \hline
    
  \end{tabular}} 
  \caption{Dataset details for sub-region classification.}
  \label{table:Dataset Details}
\end{table}


\section{Experimental Setup}\label{sec:experimental setup}
We experiment with four labelled datasets, two each from North and South Indian cities. The datasets are coded as Zone-1, Zone-2 from South India and Zone-3, Zone-4 from North India. Table \ref{table:Dataset Details} captures the details of all four datasets in terms of their size and number of classes. Each class represents a sub-region within a zone, which typically corresponds to a locality or multiple localities. The geographical area covered by a class (sub-region) varies with customer density. Thus, boundaries of these regions when drawn over a map can take extremely irregular shapes as depicted in Figure \ref{fig:last mile delivery map}. All the addresses are unique and do not contain exact duplicates. For the purpose of address classification, we do not remove cases where multiple customers ordering from the same location have written their address differently. This happens in cases where customers order from their office location or when family members order from the same house through different accounts. We set aside 20\% of rows randomly selected from each of the 4 labelled dataset as holdout test set. For all modelling approaches, we experiment with two variations of address pre-processing: Applying only the basic pre-processing step of Section \ref{sec:basic cleaning} to addresses and applying all the steps mentioned in Section \ref{sec: Address Pre-processing}. In Table \ref{table:Overall Results}, we report the classification results of all the proposed approaches. For Word2Vec with TF-IDF based approach, we train a Word2Vec model over address dataset using the Gensim\cite{rehurek_lrec} library with vectors of dimension $100$, window size of  $5$ and use it in our baseline approach mentioned in \ref{sec:tf_idf}. We train a logistic classifier with $L2$ regularization using scikit-learn\cite{scikit-learn} for $5000$ iterations 
\begin{table}[!h]
\centering
  \resizebox{\columnwidth}{!}{%
   \begin{tabular}{|c|c|c|c|c|c|}\hline
     \textbf{Modelling Approaches} & \textbf{Full Pre-Processing} & \textbf{Zone-1} & \textbf{Zone-2} & \textbf{Zone-3} & \textbf{Zone-4} \\ \hline

     Averaging Word2Vec tokens & No &0.77 & 0.79 & 0.76 & 0.80 \\ \hline 
     
     Word2Vec + TF-IDF  & No & 0.81 & 0.82 & 0.80 & 0.83 \\ \hline 
     
     Word2Vec + TF-IDF  & Yes & \textbf{0.85} & \textbf{0.84} & \textbf{0.82} & \textbf{0.85} \\ \hline 
     
     Bi-LSTM w/o pre-trained word embeddings  & No & 0.88 & 0.87 & 0.83 & 0.88 \\ \hline
     
     Bi-LSTM with  pre-trained word embeddings  & Yes & \textbf{0.89} & \textbf{0.88} & \textbf{0.84} & \textbf{0.90} \\ \hline 
     
     Pre-trained RoBERTa from hugging face  & No & 0.89 & 0.90 & 0.85 & 0.91 \\ \hline 
     
     Custom Pre-trained RoBERTa   & No & 0.88 & \textbf{0.91} & 0.84 & \textbf{0.92} \\ \hline 
     
     Custom Pre-trained RoBERTa   & Yes & \textbf{0.90} & 0.88 & \textbf{0.85} & 0.91 \\ \hline 
    
  \end{tabular}} 
\caption{Accuracy numbers for different approaches for sub-region classification task.}
\label{table:Overall Results}
\end{table}
\begin{table}[!h]
\centering
  \resizebox{\columnwidth}{!}{%
   \begin{tabular}{|c|c|c|c|}\hline
     \textbf{Dataset} & \textbf{Number of rows (addresses)} & \textbf{Perplexity} \\ \hline
   
    Combined Dataset with basic pre-processing  & 106,421 & 4.509 \\ \hline 
    
    Combined Dataset with full pre-processing & 218,434 & 5.07 \\ \hline
  \end{tabular}} 
\caption{Perplexity scores for RoBERTa pre-trained over Indian addresses for language modelling task.}
\label{table:Pre-Training Dataset Details}
\end{table}
in order to minimize cross-entropy loss.
For Bi-LSTM based approach, we experiment with two scenarios:   
\begin{enumerate}
    \item Training Bi-LSTM network by randomly initializing Embedding Matrix
    \item Training Bi-LSTM network by initializing Embedding Matrix with pre-trained word vectors obtained in Section \ref{sec:tf_idf}
\end{enumerate}
We pre-pad the tokens to obtain a uniform sequence length of $max\_length=60$ for all the addresses.  We use softmax activation in the dense layer with Adam\cite{kingma2014method} optimizer and Cross Entropy loss. The training and testing is performed individually on each of the four datasets and the number of epochs is chosen using cross-validation. For RoBERTa model, we pre-train the \textbf{`DistilRoBERTa-base'} from HuggingFace \cite{wolf2019huggingfaces}. The model is distilled from `roberta-base' checkpoint and has \textbf{6-layer}, \textbf{768-hidden}, \textbf{12-attention heads}, \textbf{82M} parameters. DistilRoBERTa-base is faster while not compromising much on performance \cite{sanh2019distilbert}. Pre-training is done on \textbf{NVIDIA TESLA P100 GPU} with a vocabulary size of \textbf{30,000} on a combined dataset of North and South Indian addresses referred as \textit{combined dataset}. We experiment with two variations of address pre-processing for pre-training the model: 
\begin{itemize}
    \item Only basic pre-processing as described in Section \ref{sec:basic cleaning}
    \item Full pre-preprocessing using all the steps mentioned in Section \ref{sec: Address Pre-processing}
\end{itemize}
Pre-training of the model is done with the above pre-processing variations and the results are present in Table \ref{table:Pre-Training Dataset Details}. The model is trained to optimize the \textbf{Masked Language Modelling} objective as mentioned in \ref{sec: BERT} for \textbf{4 epochs} with a cumulative training time of \textbf{12 hours} and a batch size of \textbf{64}. The pre-training is done using \textbf{Pytorch} \cite{NEURIPS2019_9015} framework. For sub-region classification task, we fine tune the pre-trained RoBERTa model using \textbf{\textit{`RobertaForSequenceClassification'}}\cite{wolf2019huggingfaces} and initialize it with our pre-trained RoBERTa model. Fine-tuning \textit{`RobertaForSequenceClassification'} optimizes for cross-entropy loss using AdamW \cite{kingma2014method}\cite{loshchilov2017decoupled} optimizer. 

\section{Results \& Visualization}\label{sec:results}
\begin{figure}[!h]
  \centering
  \includegraphics[width=\columnwidth]{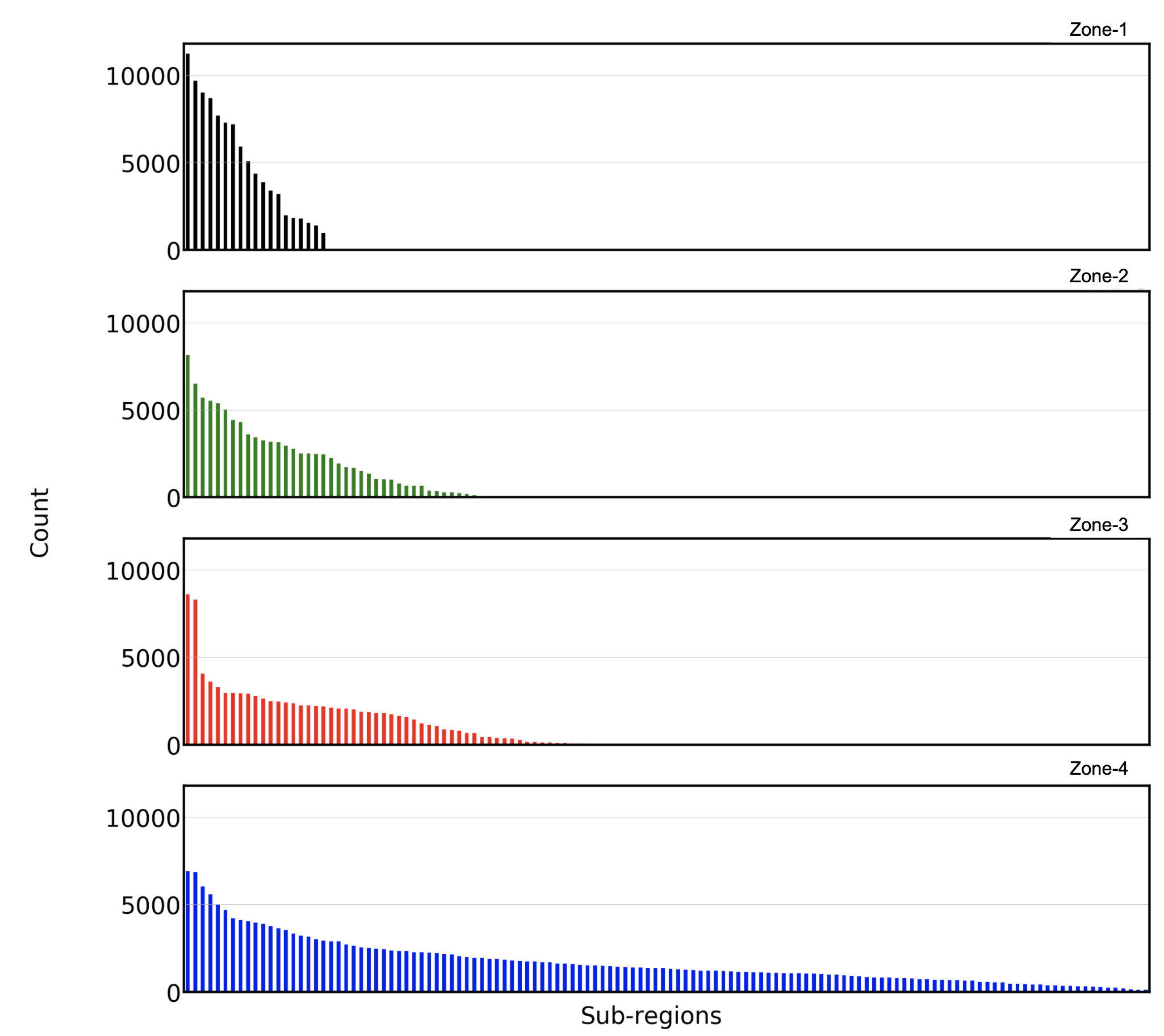}
  \caption{Class distribution of addresses for datasets mentioned in Table \ref{table:Dataset Details}.}
  \Description{}
  \label{fig:class distribution}
\end{figure}

\begin{figure}[!h]
  \centering
  \includegraphics[width=\columnwidth]{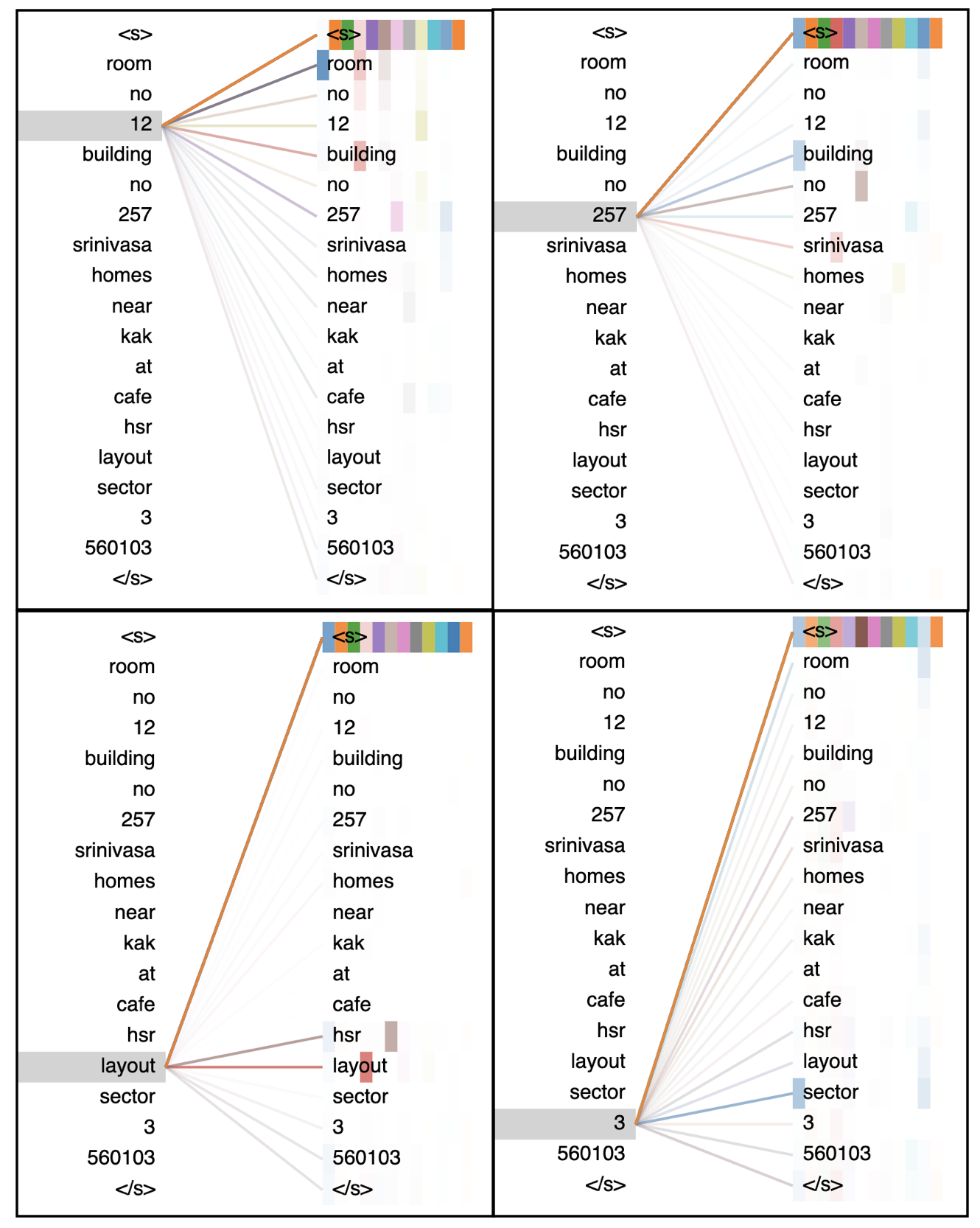}
  \caption{RoBERTa Multi-head Attention weights visualization.}
  \Description{}
  \label{fig:attention weights visualization}
\end{figure}

\begin{figure}[!h]
  \centering
  \includegraphics[width=\columnwidth]{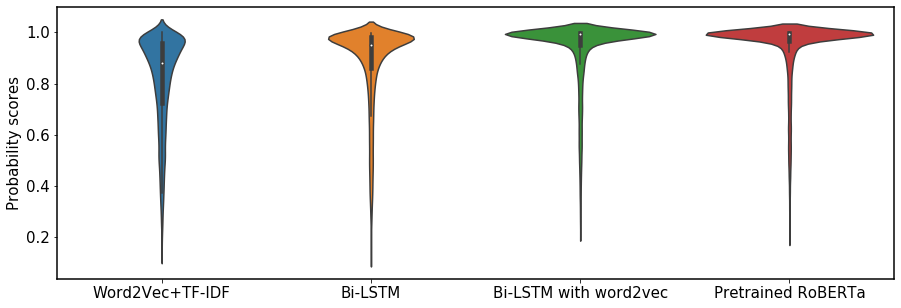}
  \caption{Violin plots of predicted class probability scores for holdout test set for Zone-1 using various approaches.}
  \Description{}
  \label{fig:violin plot}
\end{figure}

Table \ref{table:Dataset Details} lists the details of the datasets for the sub-region classification task. We use these datasets for evaluating each of the three different approaches mentioned in Section \ref{sec:approaches}. Zone-1 and Zone-2 belong to South Indian Cities and Zone-3 and Zone-4 belong to North Indian cities. One can observe that the number of addresses in Zone-1 and Zone-2 are less as compared to Zone-3 and Zone-4. The number of sub-region/classes in Zone-3 and Zone-4 are \textbf{86} and \textbf{176} respectively, which are high compared to Zone-1/Zone-2. These differences occur due the fact that different zones cater to geographical areas with different population densities. Figure \ref{fig:class distribution} shows the class distribution for the four datasets. We observe a skewed distribution since some sub-regions receive more shipments even though they are geographically small in size like office locations and large apartment complexes. Table \ref{table:Pre-Training Dataset Details} shows the combined dataset which we use for pre-training RoBERTa. We obtain a perplexity score of \textbf{4.51} for RoBERTa trained over combined dataset with basic pre-processing\footnote{basic pre-processing indicates only steps indicated in \ref{sec:basic cleaning} are applied} and with full pre-processing\footnote{full pre-processing indicates all the steps in \ref{sec: Address Pre-processing} are applied} we observe it to be \textbf{5.07}. The relatively lower perplexity scores are expected since the maximum length of addresses (sentences) after tokenization is \textbf{60} which is very less compared to the sequence lengths observed in NLP benchmark datasets like \cite{wang-etal-2018-glue}. Table \ref{table:Overall Results} shows the accuracy values of different approaches for sub-region classification task performed on each of the four holdout test sets. A \textbf{`Yes'} in the \textbf{Full Pre-Processing} column indicates that training is performed by applying full pre-processing and a \textbf{`No'} indicates only basic pre-processing steps are applied. For Word2Vec based approaches, we experiment with three scenarios: $(1)$ Simple averaging of word vectors with basic pre-processing, $(2)$ Word2Vec with TF-IDF with basic pre-processing and $(3)$ Wor2Vec with TF-IDF with full pre-processing. Among these, the third setting achieves the best accuracy scores of \textbf{0.85}, \textbf{0.84}, \textbf{0.82} and \textbf{0.85} for Zone-1, Zone-2, Zone-3 and Zone-4 respectively indicating the efficacy of TF-IDF and address pre-processing techniques. Bi-LSTM based approach with pre-trained word embeddings is able to achieve best accuracy scores of \textbf{0.89}, \textbf{0.88}, \textbf{0.84} and \textbf{0.90} which shows that using pre-trained word vectors is advantageous compared to randomly initializing word vectors in the embedding matrix. For BERT based approaches, we compare RoBERTa model pre-trained using OpenWebTextCorpus\cite{Gokaslan2019OpenWeb} with the same pre-trained using \textit{combined dataset} mentioned in Table \ref{table:Pre-Training Dataset Details}. RoBERTa pre-trained on \textit{combined dataset} of addresses with basic pre-processing is able to achieve accuracy scores of \textbf{0.88}, \textbf{0.91}, \textbf{0.84} and \textbf{0.92} respectively. Using \textit{combined dataset} with full pre-processing for pre-training, the model achieves accuracy scores of \textbf{0.90}, \textbf{0.88}, \textbf{0.85} and \textbf{0.91}. Hence, RoBERTa model pre-trained over a large address corpus for language modelling and fine-tuned for sub-region classification attains the highest accuracy scores compared to all other approaches indicating that custom pre-training over addresses is advantageous. Although the accuracy scores of Bi-LSTM and RoBERTa are comparable, there are some key differences between the two approaches. While Bi-LSTM networks are trained specifically for sub-region classification, the pre-trained RoBERTa model is only fine-tuned by training the last classification layer and hence can be used for multiple other downstream tasks using transfer learning. While we can also pre-train Bi-LSTM models as mentioned in \citet{ELMO} and \citet{ULMFit}, their performance over standard NLP benchmark datasets is significantly less as compared to BERT based models and motivates us to directly experiment with BERT based models for pre-training. BERT based models are more parallelizable owing to their non-sequential nature which is also a key factor in our choice of model for pre-training. Figure \ref{fig:attention weights visualization} shows the visualization of self attention weights for pre-trained Roberta model. We obtain these visualization using tool developed by \citet{bert-viz}. In the figure, edge density indicates the amount of weightage that is given to a particular token while optimizing for MLM loss in pre-training. We take a hypothetical example address: \textit{\textbf{`room no 12 building no 257 srinivasa homes near kakatcafe hsr layout sector 3 560103'}} for understanding the visualization. The figure at the top left shows the context in which number \textit{\textbf{`12'}} appears. From the figure it is visible that model has learned the representation in context of \textit{\textbf{`room'}} and \textit{\textbf{`building'}} which are associated with number \textit{\textbf{`12'}}. The top right figure shows the association of number \textit{\textbf{`257'}} with context tokens: \textit{\textbf{`building'}}, \textit{\textbf{`no'}} and \textit{\textbf{`srinivasa'}}. Bottom left figure indicates the association of \textit{\textbf{`layout'}} with \textit{\textbf{`hsr'}} and the one at the bottom right indicates the association of \textit{\textbf{`3'}} with \textit{\textbf{`sector'}}. We can observe from these visualizations that BERT based models incorporate the concept of self-attention by learning the embeddings for words based on the context in which they appear. From these visualizations one can find similarities in the way transformer models understand addresses and the way humans do, which is by associating different tokens to the context in which they appear. Figure \ref{fig:violin plot} shows the violin plots of probability scores assigned to the predicted classes for different approaches. We plot results for the best models in all three approaches i.e Word2Vec with TF-IDF using full pre-processing, Bi-LSTM with pre-initialized word vectors and RoBERTa model pre-trained over address corpus with full address pre-processing. One can observe that the plots become more and more skewed towards the maximum as we move from left to right indicating a decrease in entropy of predicted probability scores. This is indicative of the higher confidence with which RoBERTa model is able to classify addresses as compared to other models. 
\section{Error Analysis}\label{sec:error analysis}
In this section, we present an analysis of the cases where the approaches presented in this paper fail to yield the expected result. The address pre-processing steps mentioned in Section \ref{sec: Address Pre-processing} are heuristic methods and are prone to errors. Although these methods incorporate both edit distance threshold and phonetic matching conditions, they tend to fail in case of short-forms of tokens that customers use like \textit{\textbf{`apt'}} for \textbf{\textit{`apartment'}}, \textbf{\textit{`rd'}} for \textbf{\textit{`road'}} etc. Also the pre-processing is dictionary based token substitution and hence it fails to correct tokens when new spell variants are encountered. For the approach described in Section \ref{sec:tf_idf}, the model cannot account for sequential information in addresses. At the same time, Word2Vec approach is not able handle Out-Of-Vocabulary (OOV) tokens which poses a major hurdle with misspelled tokens in addresses. When locality names are misspelled, the tokens are simply ignored while computing the address embedding resulting in misclassification. For Bi-LSTM based approach mentioned in Section \ref{sec:bilstm}, the number of training parameters depend on the size of the tokenizer vocabulary. The tokenizer treats all OOV tokens as `UNK' (Unknown) and thus, even in this approach the OOV problem persists. Roberta model described in Section \ref{sec: BERT} is able to handle spell variations as well as out of vocabulary words using BPE encoding. In this scenario, we can set a desired vocabulary size and the OOV tokens are split into sub-tokens. When we analyze the misclassified instances of our 
\begin{savenotes}
\begin{table}[!h]
\centering
  \resizebox{\columnwidth}{!}{%
   \begin{tabular}{|c|c|c|c|}\hline
     \textbf{Type of address} & \textbf{Example} & \textbf{Probable reason} \\ \hline
    Incomplete  &\textit{`House No. XXX, Noida'} & Missing locality names \\ \hline 
    Incoherent & \textit{`Near Kormangala, Hebbal'} & Disjoint locality names\footnote{Kormangala and Hebbal are separate localities in Bangalore with no geographical overlap} \\ \hline
    Monkey typed & \textit{`dasdasdaasdad'} & Fraud/Angry customer\\ \hline
  \end{tabular}} 
\caption{Types of misclassified addresses.}
\label{table:bad addresses}
\end{table}
\end{savenotes}
RoBERTa based approach, we find that classification errors could be broadly attributed to the categories mentioned in Table \ref{table:bad addresses}. Such addresses are hard to interpret even for geo-coding APIs and human evaluators. Our approaches assign a significantly low class probability to such cases and can help in flagging those instances.

\section{Conclusion \& Future Work}
\label{sec:conclusion}
In this paper we tackled the challenging problem of understanding customer addresses in e-commerce for the Indian context. We listed errors commonly made by customers and proposed methodologies to pre-process addresses based on a combination of edit distance and phonetic algorithms. We formulated and compared different approaches based on Word2Vec, Bi-directional LSTM and RoBERTa with respect to sub-region classification task. Evaluation of the approaches is done for North and South Indian addresses on the basis of accuracy scores. We showed that pre-training RoBERTa over a large address dataset and fine-tuning it for classification outperforms other approaches on all the datasets. Pre-training Bi-LSTM based models and using them for downstream task is possible but is slow as compared to BERT variants. Recent research highlights that BERT models are faster to train and capture the context better as compared to Bi-LSTM based models resulting in state-of-the-art-performance on benchmark NLP datasets. This motivated us to use RoBERTa by pre-training it over large address dataset and subsequently fine-tuning it. As part of future work, we can experiment with different tokenization strategies like WordPiece\cite{BERT} and SentencePiece\cite{sentencePiece} for tokenizing addresses. We can also pre-train other variants of BERT and compare them based on perplexity score. Such models can generalize better in situations where labelled data is limited like address geo-coding. By framing the problem of parsing address as a language modelling task, this paper presents the first line of research using recent NLP techniques. The deep contextual address embeddings obtained from RoBERTa model can be used to solve multiple problems in the domain of Supply Chain Management.

\section{Acknowledgements}
\label{sec:acknowledgements}
The authors would like to thank Nachiappan Sundaram, Deepak Kumar N and Ahsan Mazindrani from Myntra Designs Pvt. Ltd. for their valuable inputs at critical stages of the project.

\bibliographystyle{ACM-Reference-Format}
\bibliography{sample-base.bib}
\end{document}